\documentclass[runningheads]{llncs}
\usepackage{graphicx}
\usepackage{multirow}
\usepackage{booktabs}
\usepackage{placeins}
\usepackage{bbding}

\usepackage[misc,geometry]{ifsym}
\usepackage[table,xcdraw]{xcolor}
\usepackage{caption}
\usepackage{subcaption}
\newcommand{\repeatthanks}{\textsuperscript{\thefootnote}}

\begin{document}
%
\title{EDUKG: a Heterogeneous Sustainable K-12 Educational Knowledge Graph}

%
%
\author{Bowen Zhao\inst{1}\thanks{Equal Contributions. \protect\label{X}} \and
Jiuding Sun\inst{2,3}\repeatthanks \and
Bin Xu\inst{1,3}\textsuperscript{(\Letter)} \and
Xingyu Lu\inst{3} \and
Yuchen Li\inst{3} \and
Jifan Yu\inst{3} \and
Minghui Liu\inst{3} \and
Tingjian Zhang\inst{3} \and
Qiuyang Chen\inst{3} \and
Hanming Li\inst{3} \and
Lei Hou\inst{3} \and
Juanzi Li\inst{3}}

\authorrunning{B. Zhao et al.}
\institute{Global Innovation Exchange, Tsinghua University, Beijing, China \email{\{zhaobw21,sjd22\}@mails.tsinghua.edu.cn} \and
Khoury College of Computer Sciences, Northeastern University, Boston MA, USA
\and
Department of Computer Science and Technology,\\ Tsinghua University, Beijing, China\\
\email{xubin@tsinghua.edu.cn}}

%
\let\oldmaketitle\maketitle
\renewcommand{\maketitle}{\oldmaketitle\setcounter{footnote}{0}}
\maketitle              
\begin{abstract}
Web and artificial intelligence technologies, especially semantic web and knowledge graph (KG), have recently raised significant attention in educational scenarios. Nevertheless, subject-specific KGs for K-12 education still lack sufficiency and sustainability from knowledge and data perspectives. To tackle these issues, we propose EDUKG, a heterogeneous sustainable K-12 \textbf{Edu}cational \textbf{K}nowledge \textbf{G}raph. We first design an interdisciplinary and fine-grained ontology for uniformly modeling knowledge and resource in K-12 education, where we define 635 classes, 445 object properties, and 1314 datatype properties in total. Guided by this ontology, we propose a flexible methodology for interactively extracting factual knowledge from textbooks. Furthermore, we establish a general mechanism based on our proposed generalized entity linking system for EDUKG's sustainable maintenance, which can dynamically index numerous heterogeneous resources and data with knowledge topics in EDUKG. We further evaluate EDUKG to illustrate its sufficiency, richness, and variability. We publish EDUKG with more than 252 million entities and 3.86 billion triplets. Our code and data repository is now available at https://github.com/THU-KEG/EDUKG.
\keywords{Ontology  \and Knowledge Graph \and K-12 Education}
\end{abstract}
\section{Introduction}

\emph{The object of education is to prepare the young to educate themselves throughout their lives}, as said by Robert M. Hutchins. Education, especially for K-12 children, plays a significant role in everyone's life. Intelligent education, which aims to leverage the Web and artificial intelligence (AI) technologies to improve students' learning efficiency~\cite{kuiper_web_2005,piech_deep_2015}, has always been an essential topic for researchers. In addition, the construction of educational knowledge graphs (KGs) is a fundamental research with variable downstream applications, including educational data mining~\cite{romero_educational_2020}, learning management systems~\cite{aliyu_development_2020}, question answering platforms~\cite{yang_design_2021}, dialogue systems~\cite{peng_task-oriented_2019}, etc.


Plentiful educational KGs have been proposed to help the development of computer-aided educational technologies. KnowEdu~\cite{chen_knowedu_2018} and K12Edukg~\cite{chen_automatic_2018} are constructed by extracting concepts and prerequisite rules from subject-specific textbooks. However, entities in these KGs are only course concepts without other essential educational resources for students. CKGG~\cite{shen_ckgg_2021} is proposed based on Chinese high-school-level geography education, yet they only integrate data for location entities. Meanwhile, several educational KGs are proposed based on massive online open courses (MOOCs). For instance, MOOC-KG~\cite{dang_mooc-kg_2019} and HEKG~\cite{zheng_construction_2017} are built upon open course data, yet their ontology can only represent subject-specific knowledge at a shallow level. In particular, there are only 4 and 6 defined classes in MOOC-KG and HEKG, respectively. Furthermore, although KGs built upon open courses consist of heterogeneous data, they cannot dynamically develop with growing resources. Additionally, most KGs based on MOOCs are designed for higher education instead of K-12 education.

Despite that several KGs have been proposed for educational usage, they suffer from the following limitations:

\noindent$\bullet \ \ $\textbf{Insufficient Knowledge Modeling.} Prior research pointed out that interdisciplinary teaching is beneficial for developing students' critical thinking, creativity, communication, and essential academia~\cite{jones_interdisciplinary_2010}. In the meantime, fine knowledge granularity is also beneficial for students' learning process~\cite{tiberghien_learning_1997}. Nevertheless, existing educational KGs only represent subject-specific knowledge on a coarse-grained level, lacking interdisciplinary entity relations.

\noindent$\bullet \ \ $\textbf{Sophisticated Data Curation.} Education aims to teach students with broad ability instead of just knowledge in textbooks~\cite{xu_chinas_2021}. Educational resources, including examination questions and beyond, are proved to be beneficial for fostering students' abilities through \emph{learning by doing}~\cite{bailey_working_1996,shohamy_power_1993}. Also, existing data repositories for education, such as MOOCCUBEX~\cite{yu_mooccubex_2021} leverages a concept graph to organize heterogeneous data altogether. However, existing educational KGs still lack adequate resources due to data heterogeneity.

\noindent$\bullet \ \ $\textbf{Neglected Information Growth.} Information for education is ever-growing from both knowledge and data perspectives. For knowledge, the educational reform in China is consistently changing the essential knowledge of education through time. For data, there are increasing online materials for students to learn. Nonetheless, prior educational KGs lack maintenance sustainability, i.e., the ability to capture and infuse new knowledge and resources incrementally.


\begin{figure}[ht]
    \centering
    \includegraphics[width=\textwidth]{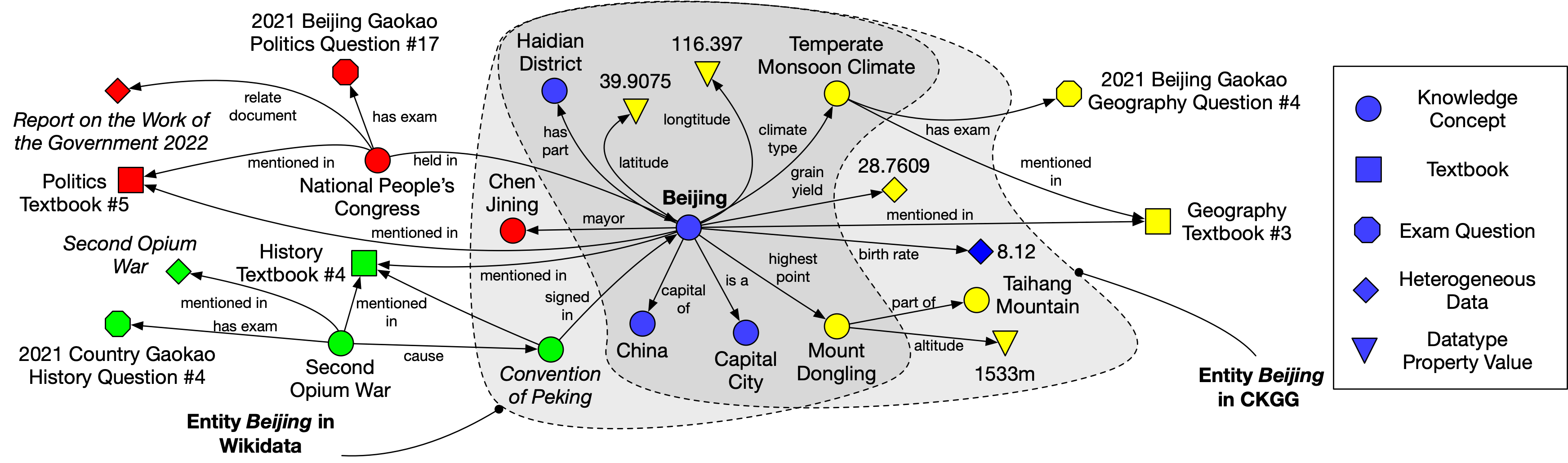}
    \caption{EDUKG's data sufficiency compared with other KGs. Blue, yellow, red, and green points refer to knowledge and resources in interdisciplinary, geography, politics, and history subjects, respectively.}
    \label{fig:beijing-in-kg}
\vspace*{-\baselineskip}
\end{figure}

To address these issues, we conclude that educational KGs should be built with an interdisciplinary schema that can represent not only knowledge but also resources. Meanwhile, towards maintaining sustainability, an educational KG should be able to grow and adapt incrementally to the change of real-world knowledge. Therefore, we propose EDUKG, a heterogeneous sustainable K-12 \textbf{Edu}cational \textbf{K}nowledge \textbf{G}raph for Chinese high-school-level education. We design an interdisciplinary fine-grained ontology that uniformly models knowledge, resources, and heterogeneous data. In total, we define 635 classes and 1759 properties in EDUKG ontology without subject boundaries. Guided by this ontology, we propose a semi-automated method for interactively acquiring knowledge from textbooks. Fig.~\ref{fig:beijing-in-kg} compares data in EDUKG with Wikidata\footnote{https://www.wikidata.org} and CKGG, indicating that EDUKG consists of most sufficient information from both knowledge and data perspectives. Additionally, for sustainably maintaining EDUKG with growing data, we propose a general mechanism to index heterogeneous online data incrementally based on our proposed entity linking technique.

\noindent\textbf{Contributions.} In general, our contributions are summarized as follows:
\begin{enumerate}
    \item An interdisciplinary, fine-grained ontology uniformly represents K-12 educational knowledge, resources, and heterogeneous data with 635 classes, 445 object properties, and 1314 datatype properties;
    \item A large-scale, heterogeneous K-12 educational KG with more than 252 million entities and 3.86 billion triplets based on the data from massive educational and external resources;
    \item A flexible and sustainable construction and maintenance mechanism empowers EDUKG to evolve dynamically, where we design guiding schema of the construction methodology as \emph{hot-swappable}, and we simultaneously monitor 32 different data sources for incrementally infusing heterogeneous data.
\end{enumerate}

\noindent\textbf{Outline.} In the following sections, we first illustrate the ontology for EDUKG in Sec.~\ref{ontology}, and we present EDUKG construction and maintenance mechanisms in Sec.~\ref{construction}. Afterward, in Sec.~\ref{quality}, we introduce essential characteristics of EDUKG to prove its sufficient qualities. In Sec.~\ref{impact-availability}, we present the impact and availability of EDUKG with its data, code, and applications. Finally, the related works are introduced in Sec.~\ref{related-work}, and we conclude our paper in Sec.~\ref{conclusion}

\section{Schema of EDUKG} \label{ontology}

\begin{figure}[ht]
\centering
\includegraphics[width=0.8\textwidth]{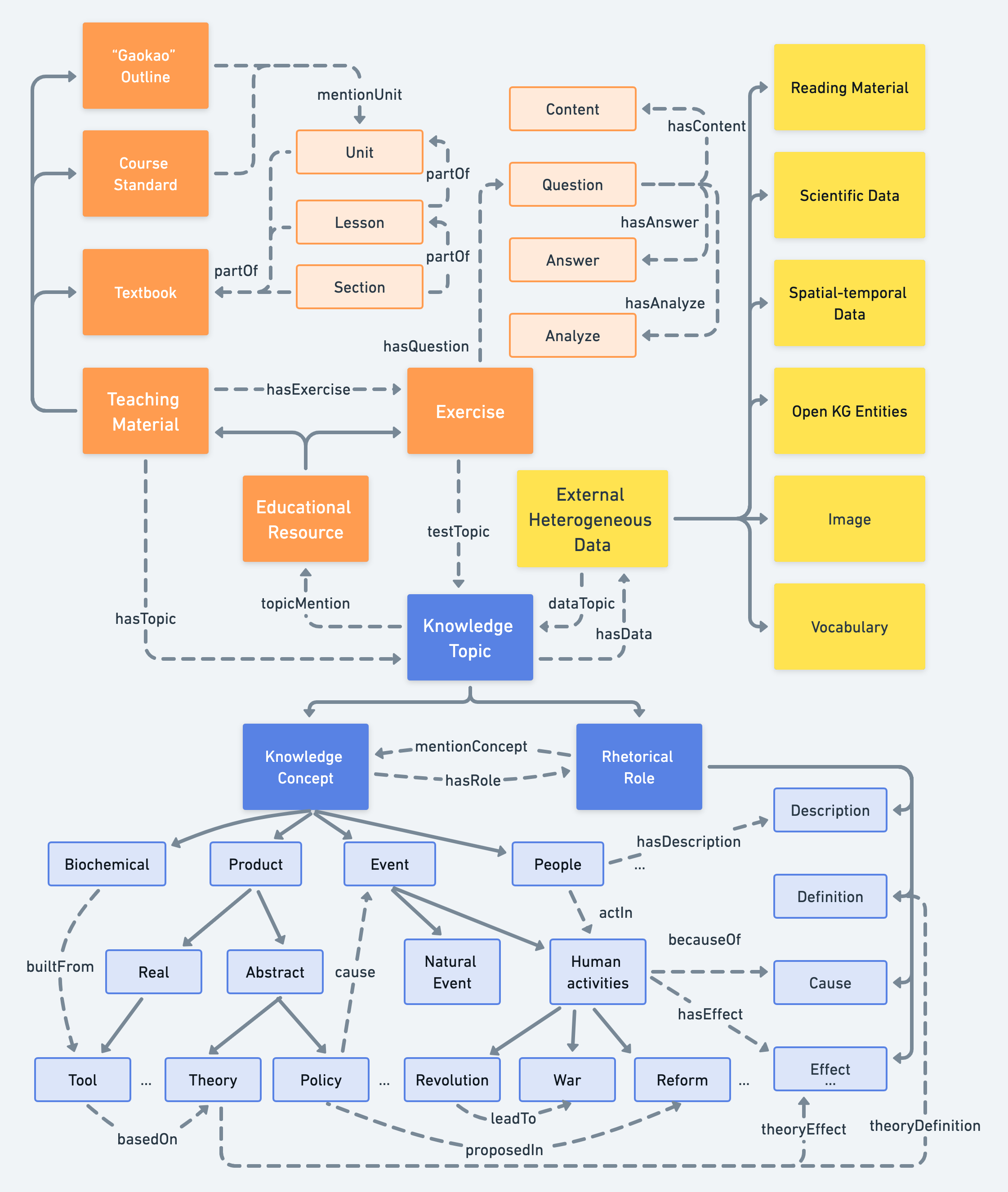}
\caption{An overview of EDUKG top-level ontology.} \label{fig1}
\vspace*{-1\baselineskip}
\end{figure}

In this section, we introduce the ontology of EDUKG, which uniformly represents knowledge, resource, and heterogeneous data. 

\subsection{Overview of EDUKG Ontology}

As shown in Fig.~\ref{fig1}, we divide EDUKG ontology into three main sections, which are ``Knowledge Topic'', ``Educational Resource'', and ``External Heterogeneous Data''.
Here we define the three top-level classes in EDUKG as follows:
\begin{itemize}
  \item ``Knowledge Topic'': essential themes in some specific subjects~\cite{ilkou_educor_2021} and their essential rhetorical roles.
  \item ``Educational Resource'': intra-curricular teaching and testing resources in K-12 education, for example, textbooks and examination exercises.
  \item ``External Heterogeneous Data'': extra-curricular resources and data give students vast approaches to learning more comprehensive knowledge.
\end{itemize}

Since EDUKG contains educational knowledge and resources, we investigate and follow multiple published knowledge and resource modeling standards. For knowledge, we reuse vocabularies from the widely-adopted RDF and RDFS schema. As for resources, we adopt the LRMI Standard\footnote{https://www.dublincore.org/specifications/dublin-core/dces}. Furthermore, we use OWL for ontology representation and reuse the schema in existing ontologies and KGs, such as schema.org\footnote{https://schema.org/}, YAGO~\cite{pellissier_tanon_yago_2020}, and Wikidata.

\subsection{Intra- and Extra-curricular Resources} \label{ontology-resource}
\begin{figure}[ht]
\centering
\includegraphics[width=\textwidth]{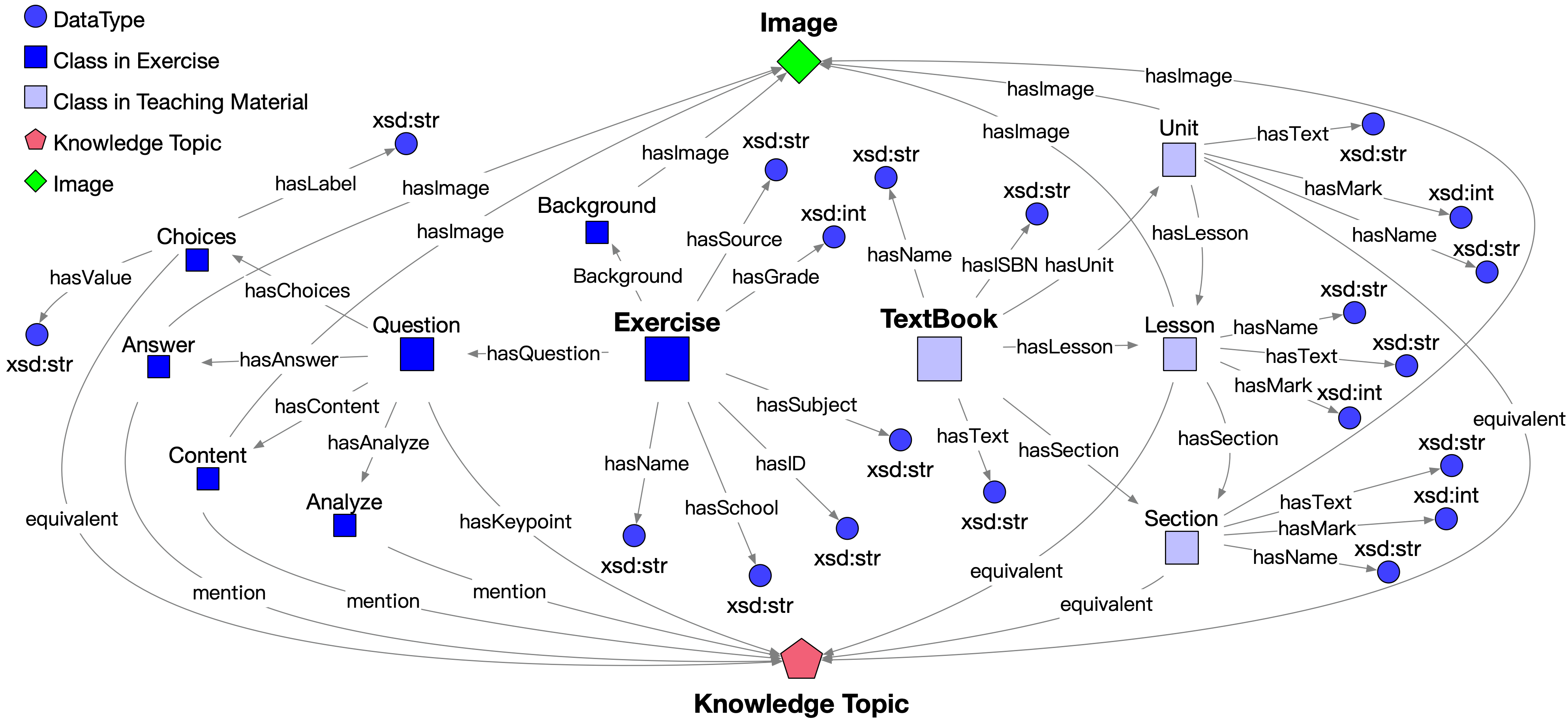}
\caption{Schema under ``Educational Resource'' class in EDUKG.} \label{fig3}
\vspace*{-1\baselineskip}
\end{figure}

We divide resources in EDUKG into two sub-classes, i.e., intra-curricular ``Educational Resource'' and extra-curricular ``External Heterogeneous Data''. Meanwhile, ``Educational Resource'' consists of two main sub-classes, ``Teaching Material'' and ``Exercise'', where the former focuses on \emph{learning to do} and the latter focuses on \emph{learning by doing}. Their detailed composition is shown in Fig.~\ref{fig3}.

\noindent\textbf{Educational Resource.} We first illustrate the two main sub-classes of ``Educational Resource''. ``Teaching Material'' represents materials delivering essential knowledge to students, i.e., \emph{learning to do}, including textbooks, learning guidance, curriculum standards, etc. Meanwhile, ``Exercise'' refers to exam questions from ``Gaokao''\footnote{also known as the National College Entrance Examination (NCEE) in China} and textbooks, which can be considered as tools to both examine and promote students' level of understanding and applying the knowledge they have learned, i.e., their ability of \emph{learning by doing}.

\noindent\textbf{External Heterogeneous Data.} To increase the coverage of extra-curricular knowledge topics mentioned or even specifically tested in ``Gaokao'', we design the ``External Heterogeneous Data'' class to represent heterogeneous data from multiple sources to enhance EDUKG. To support downstream tasks across different subjects, we gather data from multiple sources on the Web, including webpages, images, tabular data, etc. Hence, We define the sub-classes of ``External Heterogeneous Data'' according to their format, as shown in Fig.~\ref{fig1}.

\subsection{Interdisciplinary Knowledge Topic}
For knowledge modeling, we propose a fine-grained and interdisciplinary sub-ontology for EDUKG's ``Knowledge Topic'' class with two sub-classes, namely ``Knowledge Concept'' and ``Rhetorical Role''. 

\noindent\textbf{Knowledge Concept.} ``Knowledge Concept'' is a vital sub-class of ``Knowledge Topic'' for it refers to the main subjects taught during the course~\cite{yu_course_2019}, for instance, \emph{Industrial Revolution}, \emph{Bourgeoisie}, \emph{Britain}, etc. 
Since there are numerous existing ontologies, we do not need to build this sub-ontology from scratch. We adopt and modify ontologies from schema.org and YAGO as the top-level sub-ontology. For fine-grained classes on the bottom side of the hierarchy, we infuse classes and properties from several subject-specific ontologies~\cite{hu_approach_2016,siliang_drte_2018,yang_accurate_2018} proposed for K-12 education. We further enrich essential properties for these classes according to a series of popular study guides for Chinese high-school education. In all, we create 585 classes, 336 object properties, and 1177 datatype properties for EDUKG ``Knowledge Concept'' sub-ontology.

\begin{table}[ht]
\vspace*{-1\baselineskip}
\scriptsize
\centering
\caption{Essential Rhetorical Roles in EDUKG.}
\label{tab:rhetorical-role}
\begin{tabular}{p{0.12\linewidth} |p{0.4\linewidth} | p{0.45\linewidth}}
\toprule
Name         & Description                                            & Example                                                                                 \\ \midrule
Definition   & Specifically defining as truth in the context of K-12 education & Equation \textbf{is defined as} the mathematical statement consisting of an equal symbol.       \\
Process      & Processes, developments and operations      &  \textbf{Step 1.} Formulating a hypothesis.                               \\
Mechanism    & Describing mechanism and theory.                        & Fire extinguishers \textbf{work by} separating the fuel from the oxygen.                        \\
Reason        & Expressing reasons and explanations                    & The emergence of capitalism is one of the \textbf{cause} of industrial revolution.                \\
Effect       & Expressing cause and effect                            & An increase of wealth is one of the \textbf{effect} of industrial revolution.                     \\
Significance & Describing Significance                                & Carbon dioxide is an \textbf{important} greenhouse gas that helps to trap heat. \\
Condition    & Stating the condition of a proposition                  & The domain of the equation \textbf{must be} the subset of all real numbers.                       \\ \bottomrule
\end{tabular}
\end{table}

\noindent\textbf{Rhetorical Role.} \label{rhetorical-role}
Unlike open-domain KGs, for further improving knowledge granularity, knowledge topics in educational KGs should contain more than basic course concepts, i.e., named entities. The reason is that some rhetorical roles of concepts are also crucial for students during their study procedure. Rhetorical roles are defined as semantic units that segment documents into coherent units of information~\cite{malik_semantic_2021}. For example, the concept \emph{Industrial Revolution} has the rhetorical role \emph{the influence of Industrial Revolution toward Chinese society}, which is also a vital knowledge topic. Moreover, if only regarding rhetorical roles as datatype properties of knowledge concepts, some core concepts mentioned in a rhetorical role's name and content cannot be fully expressed using relations between entities in the KG. Meanwhile, using more sophisticated representations such as first-order logic to represent rhetorical roles is too difficult for users to understand. Thus, we define the ``Rhetorical Role'' class as sentences or phrases that illustrate a concept's essential properties, such as description, reason, result, significance, etc. The detailed definition and examples of some essential rhetorical roles are shown in Table~\ref{tab:rhetorical-role}.


\section{Construction and Maintenance of EDUKG} \label{construction}

\begin{figure}[ht]
\vspace*{-2\baselineskip}
    \centering
    \includegraphics[width=\textwidth]{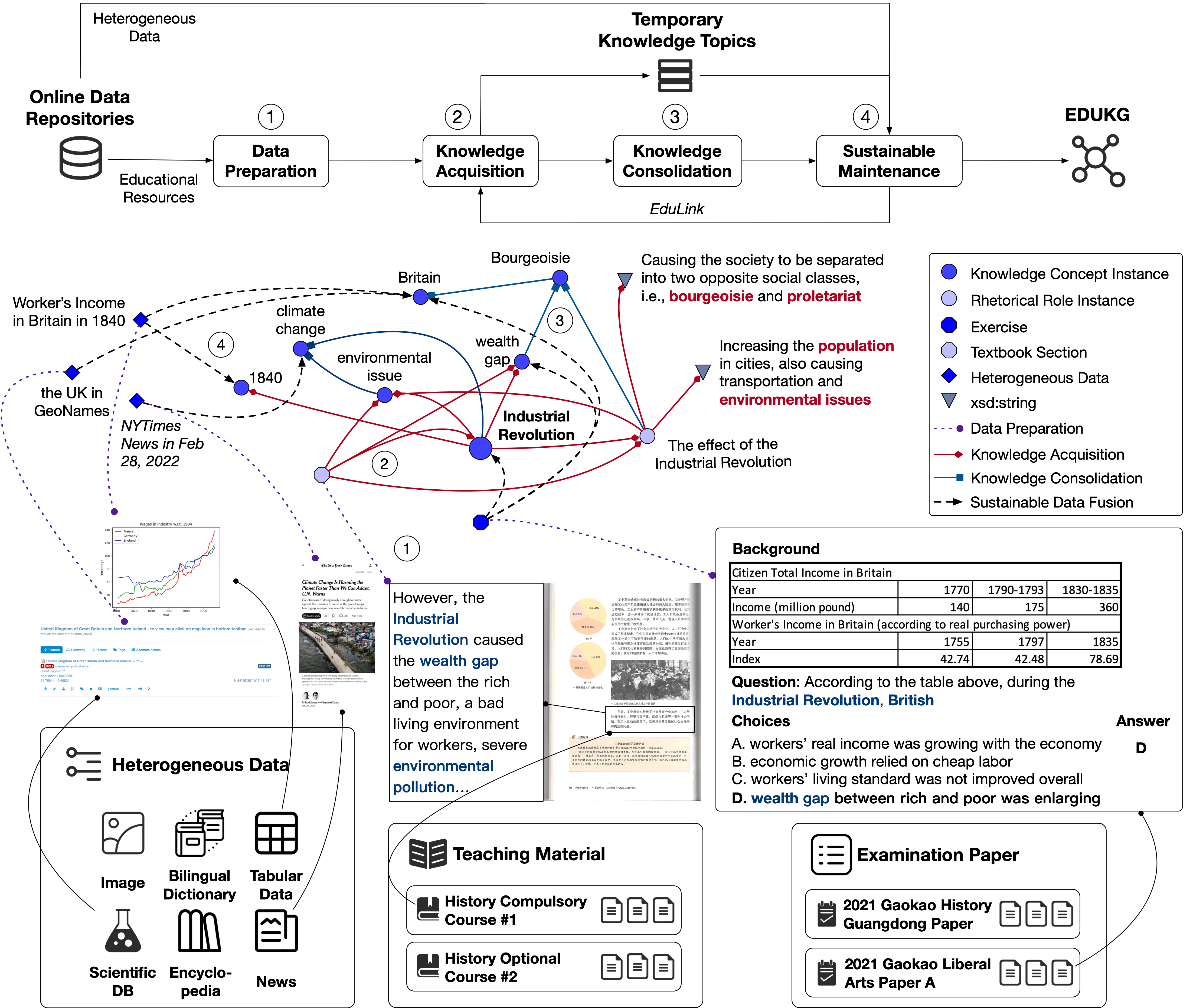}
    \caption{The construction framework of EDUKG.}
    \label{fig:framework}
\vspace*{-2\baselineskip}
\end{figure}

In this section, we introduce the EDUKG construction and maintenance framework as shown in Fig.~\ref{fig:framework}. We first introduce our data preparation procedure. We then use a semi-automated method to acquire and consolidate essential knowledge from teaching resources. Finally, we propose a sustainable EDUKG maintenance mechanism generalized from our proposed entity linking system that can consistently index massive, heterogeneous online data.

\subsection{Data Preparation} \label{hetero-data-collect}

\noindent\textbf{Teaching Material.} We first gather a complete series of textbooks for Chinese K-12 education from an online repository, including 208 books that cover 16 different subjects. Towards data adequacy, we only select 46 books from the eight main subjects (literature, mathematics, physics, chemistry, biology, history, geography, and politics) of Chinese high-school-level education since they are highly correlated with ``Gaokao'' in most regions of China. 

To increase the granularity of teaching materials, as shown in the bottom middle part of Fig.~\ref{fig:framework}, we develop an HTML-based parsing algorithm for segmenting the selected 46 textbooks into 256 units, 779 lessons, and 2371 sections in total. Afterward, we manually identify 2608 topics from \emph{Chinese national curriculum standards for high-school education} and \emph{``Gaokao'' outline} as the key topic candidates for those sections. We identify key concepts for each section according to their semantic similarity and prerequisite rule. In general, we summarize the similarity score between sections and topics as follows:
\begin{equation}
    \textrm{S}(d_i, c) = \mathcal{D}(d_i, c) \cdot \prod_{j=1}^{i-1} \mathcal{O}(d_j, c)
\end{equation}
where $d_i$ refers to the i-th section in textbooks, while $c$ is the identified concept. $\mathcal{O}(d_j, c)$ the truth function denotes whether concept $c$ is mentioned in section $d_j$, and $\mathcal{D}$ is the cosine-similarity function of TF-IDF embedding.

\noindent\textbf{Examination Papers.} For timeliness, we only gather ``Gaokao'' exam papers within the past five years. We collect data from two online examination databases and gather 302 examination papers with 6518 exercises. We categorize the exercises into five sub-classes, namely \emph{choice question}, \emph{text-filling question}, \emph{number-filling question}, \emph{question and answer}, and \emph{writing question}. Afterward, we split each exercise into four main sections, i.e., ``Background'', ``Question'', ``Answer'', and ``Analysis''. More specifically, for \emph{choice question}, there is another special section ``Choice'' defined for it as illustrated in Fig.~\ref{fig:framework}. To resolve exercise data according to the schema mentioned in Sec.~\ref{ontology-resource}, we develop a template-based method to parse and classify those 6518 exercises. Then we manually correct some wrongly parsed exercises. Additionally, we ask human annotators to manually identify the key concepts of each exercise based on its content.

\subsection{Semi-automated Knowledge Acquisition}
To construct EDUKG with high precision and broad coverage, we apply a semi-automated method for acquiring knowledge from teaching materials. We further adopt several NLP techniques to consolidate EDUKG to improve its fact coverage and completeness.

\noindent\textbf{Knowledge Acquisition from Textbooks.} \label{knowledge-acquisition}
To acquire key topics from teaching materials, we leverage both named entity recognition (NER) and entity linking methods for detecting in-text entities because NER can provide broad coverage while entity linking can ensure precision. For NER, We fine-tune the Chinese RoBERTa~\cite{cui_pre-training_2021} model on CLUENER~\cite{xu_cluener2020_2020} dataset for extracting fine-grained entities from textbooks. As for EL, we use XLink~\cite{zhang_xlink_2017} system for discovering open-domain knowledge concepts. We also use the entity linking system proposed in EDUKG (EduLink), which will be introduced in detail in Sec.~\ref{knowledge-index}. As shown in the middle part of Fig.~\ref{fig:framework}, the knowledge concepts ``Industrial Revolution'' and ``wealth gap'' are identified via the XLink system. It is worth noting that the backbone KG for EduLink is interactively changing during the human annotation. Each identified entity's confidence level is calculated as:
\begin{equation}
P(c) = S(c) + \alpha \cdot (f_{pos}(c) + f_{neg}(c))
\end{equation}
where $S(c)$ is the sum of scores calculated by NER and EL algorithms, and $f_{pos}, f_{neg}$ correspond to the frequency of positive and negative labels of the given entity, while $\alpha$ is a pre-set parameter indicates the weights of human feedback.

Moreover, for easily mining relations from given documents, we further align entities' infobox data with their potential mentions in the document as candidate knowledge triplets. Meanwhile, the entities' co-occurrence pairs are also regarded as candidates. Subsequently, we use the OpenIE API provided by NewsMiner~\cite{hou_newsminer_2015} system for jointly extracting open entities and relations from textbook documents. Finally, We merge the entities and relations with span information, and we map the predicates in extracted relations to the properties defined in EDUKG ontology by calculating similarity scores via Sentence-BERT~\cite{reimers_sentence-bert_2019} model in Text2vec Toolkit~\cite{xu_textvec}.

We invite ten annotators with outstanding scores in ``Gaokao'' to label the extracted knowledge triplet candidates from teaching materials based on our extracted knowledge triplets. We divide the labeling into two stages to tackle the error-propagation issue and reduce cost labor. The first stage focuses on entity (knowledge topic) recognition, and the second focus on triplets, i.e., relations and properties extraction. Furthermore, since there is no existing entity in EduLink at the start of the annotation procedure, we infuse several pre-built KGs based on a series of study guides in China to solve this cold-start issue. Overall, we design this ontology-guided knowledge acquisition method as \emph{hot-swappable}, where the guiding ontology is decoupled with the knowledge acquisition methods, so we can flexibly change its schema based on the latest curriculum standards in China. Our detailed knowledge acquisition task design is illustrated in the Sec.~A of appendix in our supplementary materials.

\noindent\textbf{Self-supervised Knowledge Consolidation.} \label{knowledge-consolidation}
KGs built through handcrafted or semi-automated methods may suffer from several issues from both entity and relation perspectives, including \textbf{(1) Insufficient Concept Coverage} and \textbf{(2) Incomplete Entity Relation}. Due to these issues, we leverage the following techniques to consolidate EDUKG and improve its scalability and completeness.

\noindent$\bullet \ \ $\textbf{Knowledge Concept Expansion.} We adopt the entities from Xlore~\cite{jin_xlore2_2019} to expand the concepts in EDUKG for enriching its knowledge coverage. For each entity $e$ in EDUKG, we first manually identify its equivalent entity $\hat{e}$ in Xlore. We calculate the scores of all neighbouring nodes $c$ of $\hat{e}$ as follows
\begin{equation}
\textrm{score}(c) = \textrm{Sim}(\hat{e}, c) \cdot \frac{1}{|\mathcal{N}(e)|} \sum_{n \in \mathcal{N}(e)} w_n \textrm{Sim}(n, c)
\end{equation}
where $\mathcal{N}(e)$ denotes the neighbours of entity $e$ in EDUKG, while $w_n$ indicates the weight assigned to neighbour $n$ according to its relation with $\hat{e}$. We use the cosine similarity of Sentence-BERT embedding as the similarity function $\textrm{Sim}(\cdot)$. Fig.~\ref{fig:framework} indicates that by leveraging external KG, we can infuse more knowledge concepts such as entity ``climate change'' from Wikidata.

\noindent$\bullet \ \ $\textbf{Rhetorical Roles Linking.} For the convenience of data labeling, we only ask the annotators to identify rhetorical roles as concepts' property values at first. To identify rhetorical roles, we first design templates for each sub-class of ``Rhetorical Role'' and use a template-based method to recognize them from the entities' datatype properties. After recognizing them, we leverage an entity mention detection method for linking rhetorical roles with its mentioned knowledge concepts. As shown in the middle right part of Fig.~\ref{fig:framework}, we can link the rhetorical role ``The effect of the Industrial Revolution'' with the concepts ``Bourgeoisie'' and ``environmental issues'' based on its content. In total, we recognize 18080 rhetorical roles and detect 20596 concept mentions in all rhetorical roles. 

\subsection{Sustainable Maintenance} \label{maintenance-section}

\begin{figure}[!ht]
\centering
\includegraphics[width=\textwidth]{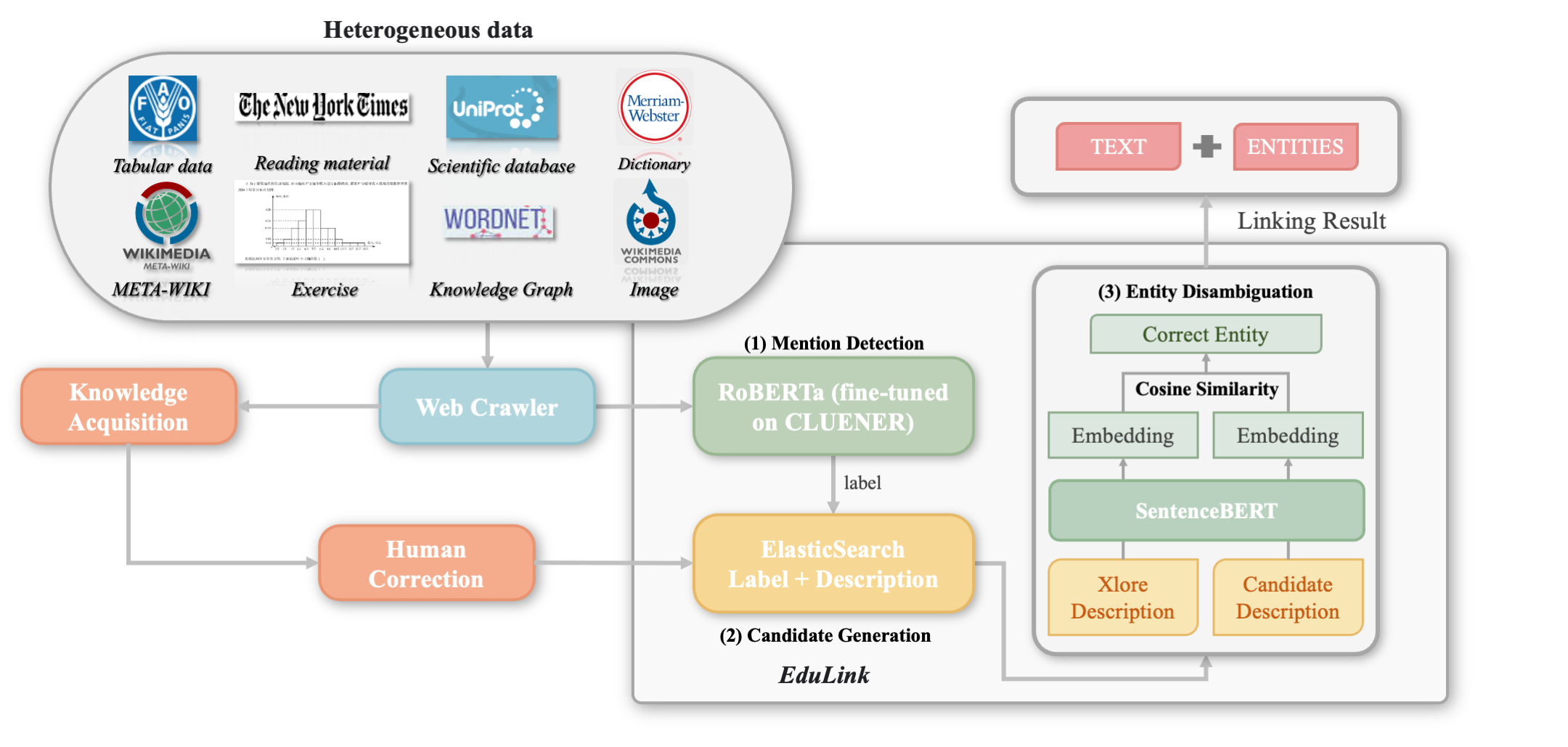}
\caption{EDUKG Sustainable Maintenance Mechanism.} \label{sustainable-maintenance}
\vspace*{-1\baselineskip}
\end{figure}

Unlike most traditional KGs built once and for all, we focus on the sustainable maintenance of our proposed EDUKG. Specifically, we regard the sustainability of EDUKG from two different aspects: (1) \textbf{Knowledge aspect}, as we have mentioned in Sec.~\ref{construction}, we design the ontology-guided knowledge acquisition method as \emph{hot-swappable}, where the schema of the knowledge acquisition can be flexibly modified based on the latest curriculum standards; (2) \textbf{Data aspect}, as shown in Fig.~\ref{sustainable-maintenance}, we propose and generalize our educational entity linking system EduLink, which leverages knowledge topics in EDUKG to dynamically index massive online data gathered through Web crawlers.

\noindent \textbf{Heterogeneous Data Compilation.}
We consistently gather massive, heterogeneous online data, including news, government policies, statistics index, lexical databases, domain-specific KGs, etc., from multiple sources via Web crawlers, and the sources of these heterogeneous data are shown in Table~\ref{tab:hetero-data}.

\begin{table}[ht]
\vspace*{-\baselineskip}
\scriptsize
\centering
\caption{Heterogeneous Data Sources for EDUKG.}
\label{tab:hetero-data}
\begin{tabular}{p{0.2\linewidth} |p{0.15\linewidth} | p{0.11\linewidth}| p{0.45\linewidth}}
\toprule
Related   Subject                                                                      & Source   & Format & Description                              \\ \midrule
\multirow{4}{*}{Biochemics}                                                            & Basechem & CSV    & Properties of chemicals                  \\
                                                                                       & PubChem  & TTL    & Chemical KG                              \\
                                                                                       & BioGRID  & TSV   & Protein and protein reaction             \\
                                                                                       & UniProt  & DAT    & Protein and gene                         \\ \midrule
\multirow{6}{*}{\begin{tabular}[c]{@{}l@{}}Comprehensive \\ Liberal Arts\end{tabular}} & FAOSTAT  & CSV    & Countries' data provided by FAO          \\
                                                                                       & IHS      & XLS    & International historical data since 1846 \\
                                                                                       & NBS      & CSV    & National and provincial data in China    \\
                                                                                       & DataBank & CSV    & Historical data in the world             \\
                                                                                       & GovNews  & JSON   & News published by Chinese government     \\
                                                                                       & GeoNames & RDF-XML    & World geographical KG                    \\ \midrule
\multirow{5}{*}{Language}                                                              & HowNet   & TXT    & Chinese sememe knowledge resource        \\
                                                                                       & NYTimes  & JSON   & New York Times news                      \\
                                                                                       & WordNet  & NT     & English electronic lexical database      \\
                                                                                       & Framster & NQ     & Integrated semantic knowledge base       \\
                                                                                       & MW       & TXT    & Merriam-Webster english dictionary       \\ \midrule
All                                                                                    & CSKG     & TSV    & Commonsense KG                           \\ \bottomrule
\end{tabular}
\vspace*{-\baselineskip}
\end{table}

We categorize the data above into three general types, which are: (1) Unstructured data, i.e., data without any pre-defined schema, including news, governmental policies, images, etc.; (2) Semi-structured data, i.e., data with limited or weak schema, such as JSON and XML files. In particular, we also regard tabular data (such as spatio-temporal geography data and statistics index) as semi-structured data, for they can only be modeled in relational databases but not in graph manners; (3) Structured data, i.e., data with specific schema or ontology, such as KGs with RDF format.

\noindent \textbf{Educational Knowledge Indexing.} \label{knowledge-index}
As illustrated in the middle left part of Fig.~\ref{fig:framework}, to use knowledge topics in EDUKG for effectively organizing gathered heterogeneous data, we propose our entity linking system, \emph{EduLink}, and generalize it as a tool for heterogeneous data indexing. Following previous research~\cite{kannan_ravi_cholan_2021}, we separate entity linking to \emph{mention detection}, \emph{candidate generation}, and \emph{entity disambiguation}. Meanwhile, we adjust several implementation details of each component for linking ``Rhetorical Role'' instances in EDUKG. We also adjust these components to support the linking for unstructured, semi-structured, and structured data. The evaluation of EduLink is presented in Sec.~B of the appendix in supplementary materials. The detailed implementation of each module is shown as follows.

\noindent$\bullet \ \ $\textbf{Mention Detection.} We leverage the aforementioned fine-tuned Chinese RoBERTa model for detecting entity mentions in plain texts. Meanwhile, since there are rhetorical roles defined as entities in EDUKG, we reuse the aforementioned template-based method in Sec.~\ref{knowledge-consolidation} for recognizing each sub-class of rhetorical roles from the given texts. Besides, only unstructured data need to be processed via the mention detection model because data with structural information are properly segmented.

\noindent$\bullet \ \ $\textbf{Candidate Generation.} As for candidate generation, to support automatic fuzzy search of entities, we load the names of EDUKG knowledge topics into \emph{ElasticSearch}\footnote{https://www.elastic.co/} along with its descriptions, and then use its searching function to generate entity candidates given the entity names detected in the step above.

\noindent$\bullet \ \ $\textbf{Entity Disambiguation.} We consider the entity disambiguation task as a sentence similarity ranking task. We calculate the similarity score between its input context and its description stored in EDUKG by the Sentence-BERT model for each detected entity. For unstructured data indexing, the input context is its original sentence or image caption; for semi-structured tabular data, the context is other properties along with their column names within the same record; for structured data, the context is the entity's description in its original KG.

\section{Evaluation of EDUKG} \label{quality}
We illustrate the quality of EDUKG in this section by presenting specific task-related characteristics. Meanwhile, we compare EDUKG with existing educational KGs, including HEKG~\cite{zheng_construction_2017}, KnowEdu~\cite{chen_knowedu_2018}, K12EduKG~\cite{chen_automatic_2018}, CKGG~\cite{shen_ckgg_2021}, MEduKG~\cite{li_medukg_2022}, and MOOC-KG~\cite{dang_mooc-kg_2019}. 
We present characteristics of EDUKG compared with existing educational KGs as shown in Table~\ref{tab:kg-stat}.

\begin{table}[ht]
\vspace*{-2\baselineskip}
\scriptsize
\centering
\caption{Characteristics comparison between existing KGs, where Pred., Rhet., Mat., Exe., Ext., and Dom. are short for predicate, rhetorical role, teaching material, exercise, external data source, subject domain, respectively.}
\label{tab:kg-stat}
\begin{tabular}{@{}lrrrr|rrrrrl@{}}
\toprule
           & \multicolumn{4}{c|}{Overall Statistics}  & \multicolumn{6}{c}{Data Sufficiency}                                                                                                                 \\ \midrule
KG         & Class & Pred. & Entity  & Triplet    & Concept & Rhet.                       & Mat.                        & Exe.                    & Ext.                        & Dom.               \\ \midrule
HEKG       & 6     & 7         & N/A     & N/A        & N/A     & \XSolidBrush & 1.2k                        & \XSolidBrush & 4                           & MOOC \\
KnowEdu    & N/A   & N/A       & N/A     & N/A        & N/A     & \XSolidBrush & \XSolidBrush & \XSolidBrush & \XSolidBrush & N/A                \\
K12EduKG   & N/A   & N/A       & N/A     & N/A        & N/A     & \XSolidBrush & \XSolidBrush & \XSolidBrush & \XSolidBrush & Mathematics        \\
CKGG       & 754   & 389       & 412k    & 1,500,000k & 412k     & \XSolidBrush & \XSolidBrush & \XSolidBrush & 20+                         & Geography          \\
MEduKG     & N/A   & N/A       & N/A     & N/A        & N/A     & \XSolidBrush & \XSolidBrush & \XSolidBrush & \XSolidBrush & N/A                \\
MOOC-KG    & 4     & 14+       & 28k     & N/A        & N/A     & \XSolidBrush & \XSolidBrush & \XSolidBrush & 4                           & MOOC \\  \midrule
Ours       & 635   & 1759      & 252,328k & 3,860,446k   & 36.79k  & 18k                         & 3k                          & 6k                          & 32                         & Interdisciplinary  \\ \bottomrule
\end{tabular}
\vspace{-1\baselineskip}
\end{table}

\begin{figure}[ht]
\vspace{-1\baselineskip}
    \centering
    \includegraphics[width=\linewidth]{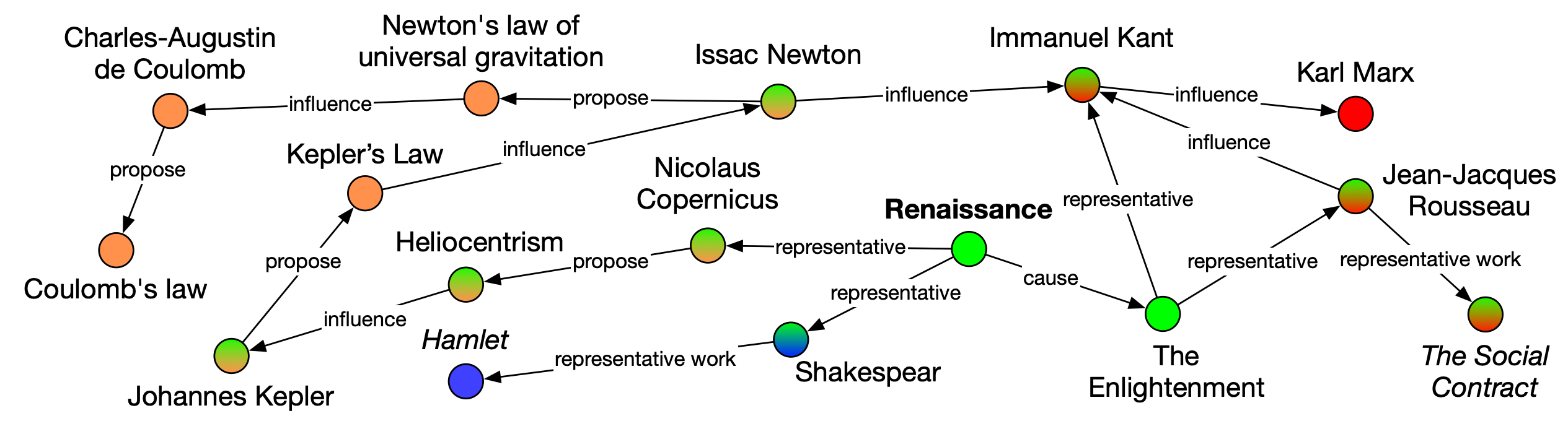}
    \caption{A case study of EDUKG interdisciplinary concepts, where green, blue, orange, and red points refer to subjects of history, literature, physics, and politics, respectively. Points with multiple colors refer to interdisciplinary concepts.}
    \label{fig:interdisciplinary-concepts}
\vspace{-1\baselineskip}
\end{figure}

\noindent\textbf{Knowledge Sufficiency.} In Table~\ref{tab:kg-stat}, although CKGG has more classes and triplets than EDUKG, its original paper claimed that there are 655 classes and 353 properties in CKGG which are not populated, meaning that EDUKG still achieves better knowledge granularity. Besides, entities in CKGG are mostly locations, and triplets in CKGG are almost datatype properties of locations, lacking data variability. Thus, Compared with existing educational KGs, EDUKG achieves remarkably better knowledge sufficiency from the following aspects:

\noindent$\bullet \ \ $\textbf{Interdisciplinary Knowledge Modeling.} EDUKG contains rich interdisciplinary knowledge topics and relations extracted from textbooks. On average, each extracted knowledge topic occurs in 1.70 different textbooks. For example, the concept \emph{Leonardo Da Vinci}, who is a great artist, scientist, engineer, and theorist, is essential in subjects like mathematics, physics, and history. Furthermore, we show a case study in Fig.~\ref{fig:interdisciplinary-concepts}, where we illustrate the sufficiency and practicality of interdisciplinary knowledge toward education.

\begin{figure}[ht]
     \centering
     \begin{subfigure}[b]{0.48\textwidth}
         \centering
         \includegraphics[width=\linewidth]{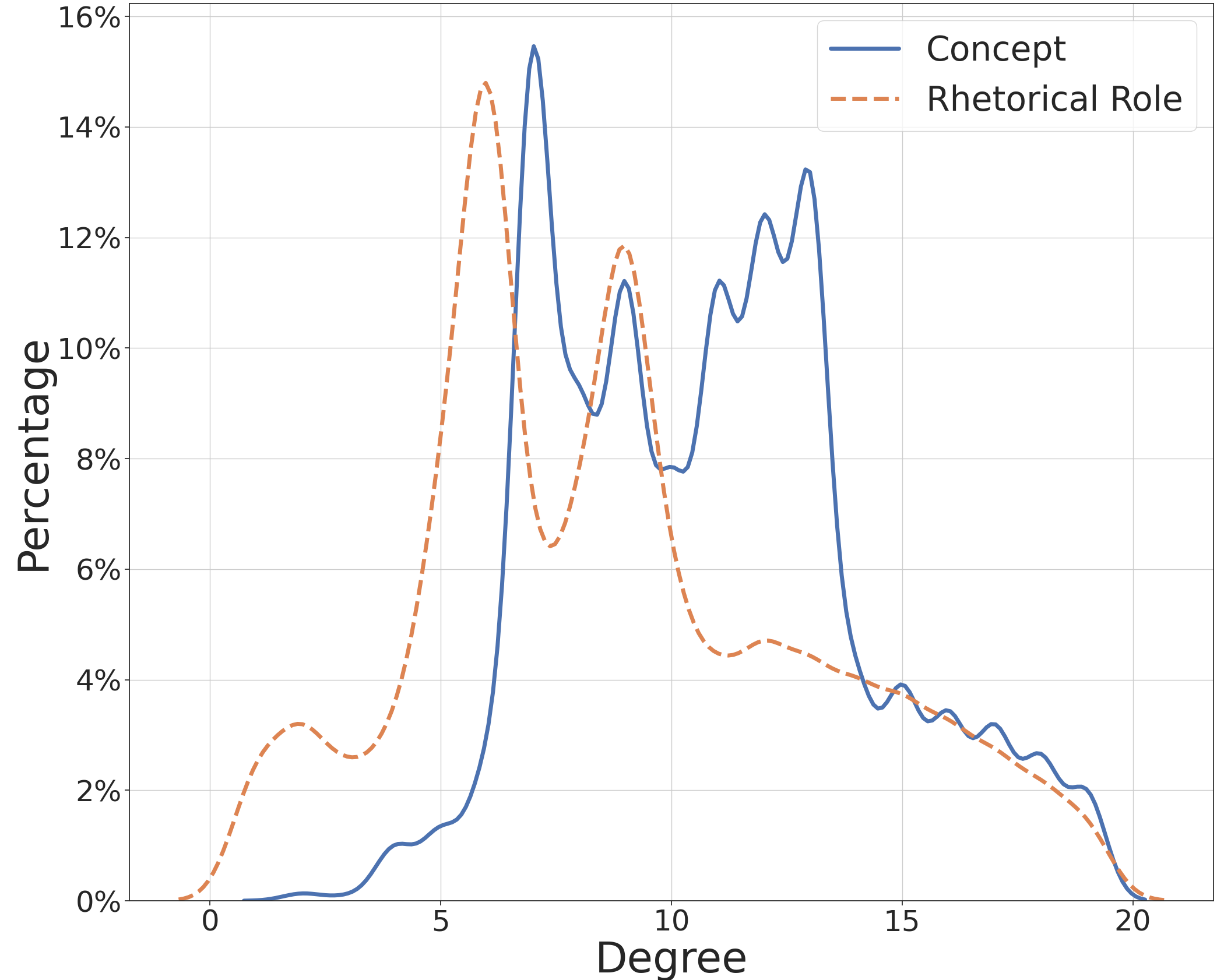}
         \caption{Topic degree.}
         \label{fig:degree-distribution}
     \end{subfigure}
     \hfill
     \begin{subfigure}[b]{0.48\textwidth}
         \centering
         \includegraphics[width=\linewidth]{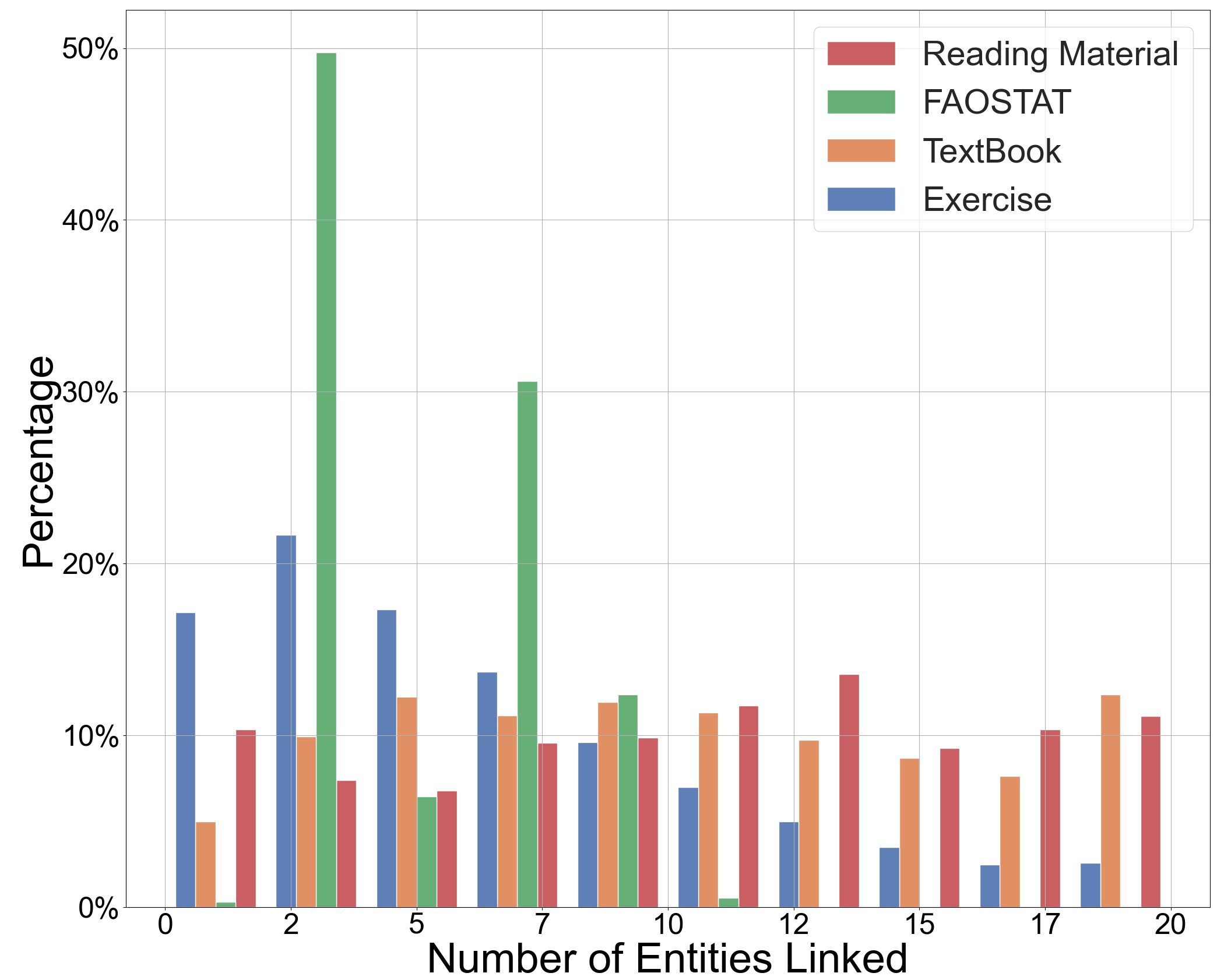}
         \caption{Heterogeneous data.}
         \label{fig:heterogeneous-distribution}
     \end{subfigure}
        \caption{The detailed characteristics for knowledge and resources in EDUKG.}
        \label{fig:characteristics}
\vspace*{-1\baselineskip}
\end{figure}

\noindent$\bullet \ \ $\textbf{Fine-grained Knowledge Topics.} We organize extracted knowledge concepts based on our proposed fine-grained sub-ontology under the ``Knowledge Concept'' class. The average instances' depth on the class hierarchy of ``Knowledge Concept'' sub-ontology is 3.87, while 48.39\% of the instances of ``Knowledge Concept'' are classified under the leaf type of its class hierarchy, which addresses the fine granularity of extracted concepts. Also, as mentioned in Sec.~\ref{rhetorical-role}, to our knowledge, we are the first to infuse rhetorical roles as entities in KG to improve the knowledge granularity. Each rhetorical role, on average, mentions 1.14 knowledge concepts in EDUKG, showing that infusing rhetorical roles in KG can densify our KG effectively. The distribution of each knowledge concept and each rhetorical role's in- and out-degree are shown in Fig.~\ref{fig:degree-distribution}, presenting the high density of EDUKG's knowledge topics.

\noindent\textbf{Resource Richness.} As shown in Table~\ref{tab:kg-stat}, there are only a few existing educational KGs consisting of teaching materials or other educational-related resources. EDUKG includes 256 units, 779 lessons, and 2371 sections segmented from 46 textbooks of eight main subjects in Chinese high-school-level education. Moreover, each section has 22.17 related knowledge topics on average, indicating the high cohesion between educational resources and knowledge in EDUKG. Furthermore, there are 6518 exercises in EDUKG, which are further parsed into 10602 questions in total. Each exercise, on average, is linked with 1.66 knowledge topics, where 65.29\% among them are not explicitly mentioned yet essential for the problems' solving. 

\noindent\textbf{Data Variability.} Since EDUKG is a heterogeneous educational KG, data variability plays an essential role in EDUKG's data quality. As mentioned in Sec.~\ref{knowledge-index}, we leverage our entity linking system for indexing online heterogeneous data. In total, we convert data from 32 sources into more than 250 million entities, where on average, each entity corresponds to 4.80 knowledge topics in EDUKG. The detailed distribution of related knowledge topics for some sub-classes of ``External Heterogeneous Data'' is shown in Fig.~\ref{fig:heterogeneous-distribution}. In addition, we also infuse image data in EDUKG. On average, each textbook section is linked with 3.11 images, and each exercise is linked with 1.36 images, illustrating our data heterogeneity.

\noindent\textbf{Supported Tasks.} Prior KGs for education suffer from the lack of sufficient knowledge and heterogeneous resources and data. Thus, they could not fully support variable downstream educational tasks. However, EDUKG can be regarded not only as a KG but also as a data cube for numerous educational-related usages, including learning management system development, intelligent tutoring system research, educational data mining exploration, etc. We show a question answering demo platform in Sec.~C of the appendix in supplementary materials.


\section{Impact and Availability} \label{impact-availability}
In this section, we first present the availability of EDUKG, including its code, data, and applications. Afterward, we highlight the impact of EDUKG on research and society, along with its beneficial groups.

EDUKG is published under the terms of the Creative Commons Attribution-NonCommercial-ShareAlike 4.0 International License. EDUKG ontology is available under persistent URI in w3id\footnote{https://w3id.org/edukg/ontology/}, and the RDF dump of EDUKG is available at our Github homepage\footnote{https://github.com/THU-KEG/EDUKG}. All data correspond to knowledge, resources, and heterogeneous data is free to download. The ontology and sampled data of EDUKG can be found in supplementary materials. Additionally, our toolkits for constructing and sustainably maintaining EDUKG are also open-sourced. Furthermore, we build an open platform\footnote{http://open.edukg.cn/home} upon EDUKG, where users can efficiently browse massive K-12 educational knowledge and resources. We also provide vast kinds of open API, including knowledge searching, entity linking, question answering, etc., on our open platform for the usage of developers and researchers.

\noindent\textbf{Educational Research Development.} EDUKG can provide essential knowledge extracted from textbooks in a fine-grained manner, which is essential for researchers to explore cutting-edge educational technologies. Meanwhile, these data are also valid for developers to build downstream educational applications.

\noindent\textbf{Educational Equity Promotion.} Students nowadays from different regions still suffer from regional education inequality problem~\cite{hannum_geography_2006}. With the help of the EDUKG-based online learning platform, students can access comprehensive learning resources and conduct self-learning efficiently, which effectively ceases the education inequality issue. 

\noindent\textbf{Beneficial Groups.} EDUKG aims to support the development of variable beneficial groups, including (1) students, parents, and teachers with limited learning resources, (2) educational application developers with insufficient data, and (3) researchers of educational-AI fields to access abundant knowledge and resources from our platform efficiently.

\section{Related Work} \label{related-work}
KnowEdu~\cite{chen_knowedu_2018} is a KG construction system extracting entities and mining prerequisite rules from K-12 teaching materials. Meanwhile, K12EduKG~\cite{chen_automatic_2018} leverages existing NLP techniques, such as NER, for extracting key educational concepts from K-12 mathematical textbooks. It also proposes an association rule mining algorithm to identify prerequisite rules between entities. Besides, several educational KGs are proposed based on MOOCs. For example, MOOC-KG~\cite{dang_mooc-kg_2019} is constructed based on MOOC platforms, where numerous learning materials are integrated. Besides, HEKG~\cite{zheng_construction_2017} uses entity detection and relation extraction techniques based on a simple schema with the crawled information on MOOC websites. In addition, CKGG~\cite{shen_ckgg_2021} is claimed as a KG for Chinese high-school-level geography education. However, it should be regarded as a KG for geography rather than educational usage because it integrates massive professional geography data beyond high-school-level education. 

\section{Conclusion and Future Work} \label{conclusion}
This paper proposes EDUKG, a heterogeneous sustainable K-12 educational KG based on Chinese high-school education with more than 252 million entities and 3.86 billion triplets. To the best of our knowledge, EDUKG is the first large-scale interdisciplinary KG for K-12 education. Meanwhile, we design and propose a series of toolkits for sustainable data maintenance to dynamically collect and extract knowledge and resources from massive online data. 

As for future work, there are several approaches for improving the data sufficiency of EDUKG. A substantial improvement is to add multilingual and multimodal data in EDUKG. 
Meanwhile, for adaptive learning tasks, the student behavioral data is essential for downstream tasks. We would further conduct user experimentation based on EDUKG for broader task support. Moreover, our future work will also focus on developing variable downstream applications. For instance, we are working on several applications for real high-school education scenarios based on educational data mining. EDUKG can also provide rich knowledge and data for question answering platform development. We hope that EDUKG can be beneficial to the development of educational technologies.

%
%
%
\bibliographystyle{splncs04}
\bibliography{citations}

\end{document}


\renewcommand\thesection{\Alph{section}}

\section{Interactive Knowledge Acquisition}
We design the task of knowledge acquisition from textbooks in an interactive manner. The candidate entities and relations are first extracted by named entity recognition (NER), entity linking, and open information extraction (OpenIE). Afterward, the human annotation will provide feedback for the back-end algorithm to adjust each candidate's weight according to the human labeling result. In order to reduce the cost of labor while improving the human annotation accuracy, we split the whole knowledge acquisition task into two separate procedures, where the first one is to label entities and the second one is to label relations and properties. Furthermore, separately labeling entities and triplets will effectively cease the error propagation problem.

\subsection{Entity Recognition Labeling}
We design the entity recognition interface as shown in Fig.~\ref{fig:entity-recognition}, where the left side is the input plain texts, and the right side is the pre-extracted candidate entities. Entities to be extracted include named entities, encyclopedic entries, essential knowledge concepts, and rhetorical roles in K-12 education. 

\begin{figure}[htbp]
    \centering
    \includegraphics[width=\linewidth]{figures/Entity Recognition Labeling.png}
    \caption{Interactive knowledge topic recognition from textbooks.}
    \label{fig:entity-recognition}
\end{figure}

More specifically, during the labeling procedure, the annotators will also label each extracted concept's correlated class in EDUKG ontology and its equivalent entity in Xlore. By labeling each entity's class, we efficiently can organize knowledge topics of EDUKG in a fine-grained manner. Meanwhile, assigning each entity with its equivalents in Xlore can disambiguate some entities with the same label. For example, the concept ``capacitance'' may refer to a physical quantity or a real-world electronic component. By leveraging external knowledge, the labeled entities can support the evaluation of variable downstream KG-based tasks, such as entity linking, entity alignment, etc.

\subsection{Triplet Extraction Labeling}
As shown in Fig.~\ref{fig:triplet-extraction}, unlike traditional relation or property extraction labeling, we leverage multiple existing knowledge from external KG to support the human annotation process. In particular, inspired by distant supervision methodology, we leverage the entity relations and properties from its infobox in Xlore and EDUKG to recognize whether there are potential triplets that occurred in the given document. Besides, entities that co-occurrence in a single sentence are also considered candidate triplets, yet the relation type needs to be identified by human annotators. Furthermore, we leverage the Sentence-BERT model to \emph{canonicalize} the triplets acquired by OpenIE, where the predicates are mapped into the properties defined in EDUKG ontology according to the cosine similarity of their Sentence-BERT embeddings. At the same time, the tail description is considered the entity's datatype property or rhetorical role.

\begin{figure}[htbp]
    \centering
    \includegraphics[width=\linewidth]{figures/Triplet Extraction Labeling.png}
    \caption{Interactive triplet extraction interface.}
    \label{fig:triplet-extraction}
\end{figure}

Our knowledge acquisition labeling platform significantly reduces the labor cost of human annotation. Traditionally, in relation extraction tasks, human annotators are asked to extract the head entity, the predicate, and the tail entity for each possible triplet from scratch. We have conducted a user experiment by using the traditional methodology, where we ask ten students to annotate knowledge triplets based on the given document without candidate generation. The result shows that, on overage, each annotator can only label 4.98 sections per hour. However, each annotator can label 12-15 sections per hour using our interactive knowledge acquisition tool, which significantly increases the annotation efficiency.

\section{Entity Linking Evaluation}
In Sec.~3.3, we describe the entity linking system, \emph{EduLink}, that we use to infuse massive, heterogeneous online data. We further conduct several experiments to analyze the performance of the EduLink system. As shown in Table.~\ref{tab:entity-linking-result}, we ask several human annotators to identify mentioned knowledge topics of EDUKG in textbooks from three different subjects (physics, biology, and geography), and we compare the human-labeled result with the output of EduLink.

\begin{table}[htbp]
\centering
\caption{Entity linking performance compared with golden truth.}
\label{tab:entity-linking-result}
\begin{tabular}{@{}lrrr@{}}
\toprule
Subject   & Recall & Precision & F1 \\ \midrule
Physics   & 81.01  & 87.88     & 85.14    \\
Biology   & 86.48  & 77.01     & 81.47    \\
Geography & 66.38  & 90.8      & 76.79    \\ \bottomrule
\end{tabular}
\end{table}

Table~\ref{tab:entity-linking-result} indicates that our entity linking system can achieve a reasonable result in science and technology subjects (physics and biology). However, the recall of EduLink in geography is significantly worse than in the other two subjects. The reason might be that there exist numerous location entities in geography, yet our entity linking system focuses more on encyclopedic knowledge instead of subject-specific entities, for example, the names of specific towns and villages. 

Furthermore, we find that among the three main modules of EduLink, namely mention detection, candidate generation, and entity disambiguation, the candidate generation module may produce the most obvious deviation. The reason is that we only leverage the existing searching function provided by ElasticSearch without any further improvement based on educational scenarios. Because that ElasticSearch will segment Chinese sentences at the character level, some words, after segmenting, might produce mis-interpretation. As a result, in the future, we will continue to improve this module with our implementation based on our prior knowledge in educational settings. Our entity linking API is online available, where its documentation can be found on our open platform\footnote{http://open.edukg.cn/help}. We hope that the EduLink system will benefit broader knowledge fusion tasks in K-12 educational scenarios.

\section{Question Answering Demo}
To support intelligent tutoring systems, we also develop a question answering (QA) demo platform that can provide essential factual knowledge in K-12 education based on EDUKG. Specifically, we leverage a hybrid methodology that infuses a knowledge-based QA algorithm and a text-based QA method. We present two detailed case studies as follows.

In Fig.~\ref{fig:kbqa-case}, we present a case study with the topic ``the French Revolution''. Given the query question \emph{What is the starting time of the French Revolution}, the QA system will use a pre-defined template to recognize the corresponding predicate defined in EDUKG ontology. In this case, the word ``starting time'' is identified as the keyword of our defined datatype property. Afterward, the QA system will convert the input question into a SPARQL query based on the matched template to fetch the factual knowledge stored in EDUKG.

Fig.~\ref{fig:qa-case} illustrates another case based on the topic ``Newton's first law of motion'', where we focus on questions mainly based on text. Traditionally, because there are no rhetorical roles defined in KGs, a question like \emph{what is the content of ``Newton's first law of motion''} is difficult for the QA system to convert into a SPARQL query, for the word ``content'' might be missing in the KG ontology. However, because we define rhetorical roles as entities in EDUKG, words such as ``content'' are regarded as keywords for descriptive rhetorical roles. Thus, without searching the whole KG based on information retrieval methods, the system only needs to convert this question into a SPARQL query for rhetorical roles for generating the answer. 

\begin{figure}[ht]
     \centering
     \begin{subfigure}[b]{0.48\textwidth}
         \centering
         \includegraphics[width=\linewidth]{figures/EDUKG KBQA Case Study.png}
         \caption{QA case study in history.}
         \label{fig:kbqa-case}
     \end{subfigure}
     \hfill
     \begin{subfigure}[b]{0.48\textwidth}
         \centering
         \includegraphics[width=\linewidth]{figures/EDUKG QA Case Study.png}
         \caption{QA case study in physics.}
         \label{fig:qa-case}
     \end{subfigure}
        \caption{The detailed characteristics for knowledge and resources in EDUKG.}
        \label{fig:qa-case-study}
\end{figure}

Our QA demo system is now online available\footnote{http://edukg.cn/qa}. In general, as the development of NLP is still in progress, our future work on QA applications will focus on leveraging the sequence-to-sequence model for semantic parsing, i.e., analyzing the input question and converting it into a corresponding SPARQL query for fetching the factual knowledge. The definition of instances of class ``Rhetorical Role'' in EDUKG can be beneficial to the unification of knowledge-based QA and text-based QA in K-12 educational scenarios. As illustrated in our two case studies, we believe that EDUKG can provide a solid support for the research and development of intelligent tutoring systems.

%
%
%